\documentclass[conference]{IEEEtran}
\IEEEoverridecommandlockouts
\usepackage{cite}
\usepackage{amsmath,amssymb,amsfonts}
\usepackage{algorithmic}
\usepackage{graphicx}
\usepackage{textcomp}
\usepackage{subcaption}
\usepackage{xcolor}
\usepackage[pdfstartview=XYZ,
bookmarks=true,
colorlinks=true,
linkcolor=blue,
urlcolor=blue,
citecolor=blue,
pdftex,
bookmarks=true,
linktocpage=true, 
hyperindex=true
]{hyperref}
\usepackage{orcidlink}
\def\BibTeX{{\rm B\kern-.05em{\sc i\kern-.025em b}\kern-.08em
    T\kern-.1667em\lower.7ex\hbox{E}\kern-.125emX}}
\begin{document}

\title{Position Paper: Building Trust in Synthetic Data for Clinical AI}


\author{Krishan Agyakari Raja Babu$^{1*}$ \orcidlink{0000-0001-9362-2420} , Supriti Mulay$^{1}$, Om Prabhu$^{2}$ and Mohanasankar Sivaprakasam$^{1}$ \\
$^{1}$Indian Institute of Technology Madras, Chennai, India 600036 \\
$^{2}$All India Institute of Medical Sciences, New Delhi, India 110029 \\
$^{*}$Correspondence: ee22s042@smail.iitm.ac.in
}

\maketitle

\begin{abstract}
Deep generative models and synthetic medical data have shown significant promise in addressing key challenges in healthcare, such as privacy concerns, data bias, and the scarcity of realistic datasets. While research in this area has grown rapidly and demonstrated substantial theoretical potential, its practical adoption in clinical settings remains limited. Despite the benefits synthetic data offers, questions surrounding its reliability and credibility persist, leading to a lack of trust among clinicians. This position paper argues that fostering trust in synthetic medical data is crucial for its clinical adoption. It aims to spark a discussion on the viability of synthetic medical data in clinical practice, particularly in
the context of current advancements in AI. We present empirical evidence from brain tumor segmentation to demonstrate that the quality, diversity, and proportion of synthetic data directly impact trust in clinical AI models. Our findings provide insights to improve the deployment and acceptance of synthetic data-driven AI systems in real-world clinical workflows.
\end{abstract}


\section{Introduction}
In recent years, the use of artificial intelligence (AI) in healthcare has sparked both excitement and skepticism. On one side, advocates of synthetic medical data highlight its promise in overcoming critical challenges such as data scarcity, privacy concerns, and class imbalance. By generating artificial datasets that replicate real-world data, synthetic data offers a way to train AI models without using sensitive patient information. This ability to produce vast, diverse datasets could revolutionize the way we approach medical AI, potentially leading to more robust models and reducing biases that often plague real-world data. Proponents argue that synthetic data is not just a valuable tool but a game-changer in advancing healthcare AI systems.

However, on the opposing side, there are significant concerns. Despite the theoretical advantages of synthetic data, its practical adoption in clinical settings remains slow. The key obstacle? Trust. Trust is essential when it comes to AI in healthcare, as clinicians are tasked with making life-altering decisions based on the output of AI models. They need to be confident that these models are accurate, fair, and reliable. The question is, how can we expect clinicians to trust AI systems that rely on synthetic data, when there are still doubts about its credibility and real-world applicability? Can synthetic data truly mirror the complexities of human health, or is it simply an approximation that falls short when faced with diverse patient populations and unique clinical scenarios?

This debate centers around one critical point: fostering trust in synthetic medical data is essential for its adoption in clinical practice. Without trust, no matter how advanced the models or promising the data may be, AI systems will struggle to gain a foothold in real-world healthcare environments. It’s not enough to prove that synthetic data works in controlled experiments; we need to demonstrate that it can be trusted to function reliably across a broad range of clinical situations.

In this position paper, we argue that synthetic medical data can earn this trust, but only if we pay close attention to its quality, diversity, and proportion. Through empirical studies on brain tumor segmentation, we show that these factors directly influence how clinicians perceive the reliability and performance of AI models. Our findings provide critical insights into how we can bridge the gap between the potential of synthetic data and its real-world adoption, offering a framework for the trustworthy integration of synthetic data in clinical workflows.

\section{Related Works}
The use of synthetic data in healthcare has gained significant attention as a potential solution to challenges such as data scarcity, privacy concerns, and class imbalance. \cite{c1} highlight the growing role of synthetic data in healthcare, discussing its innovation, applications, and how it addresses privacy concerns in medical data sharing and usage . This idea is further explored by \cite{c2}, who provide a narrative review on the utility of synthetic data in healthcare, stressing its potential to augment training datasets while ensuring patient confidentiality . Additionally, \cite{c3} provide a comprehensive overview of how synthetic data can support AI systems by simulating diverse real-world scenarios, thereby enhancing the generalizability of machine learning models in medical applications .

Recent advancements in generative models, especially Generative Adversarial Networks (GANs) and diffusion models, have shown promise in generating high-quality synthetic medical images. The study by \cite{c9} compares the performance of GANs and diffusion models in generating synthetic MR images for brain tumor segmentation, demonstrating the potential of these models in medical imaging tasks . In particular, \cite{c6} explore the use of conditional diffusion models for 3D brain MRI synthesis, emphasizing how such models can generate realistic images that can be used to augment training datasets for segmentation tasks . These works underline the growing importance of synthetic image generation in improving AI performance for medical tasks like brain tumor detection.

Trust in AI models is a crucial factor for their successful adoption in healthcare, especially when synthetic data is involved. \cite{c7} discuss the importance of building trustworthy AI systems in healthcare, emphasizing principles such as fairness, transparency, and explainability . These principles are essential when using synthetic data, as clinicians must be assured that the generated data closely mirrors real-world clinical scenarios and that the AI models can generalize well to new, unseen cases. \cite{c10} further explore the challenges associated with trust in medical AI, noting that issues like data provenance, explainability, and biases in synthetic data hinder the widespread adoption of AI in clinical settings .

While synthetic data offers many advantages, it is not without its challenges. One such challenge is simplicity bias, where models trained on synthetic data fail to generalize effectively to real-world data due to inherent differences in the distributions of synthetic and real-world data. \cite{c8} examine this issue in their work on medical data augmentation, pointing out that synthetic datasets may oversimplify complex patterns, leading to models that underperform in real clinical settings . Moreover, the limited diversity in synthetic datasets can introduce biases that reduce the trustworthiness of AI models trained on these data. Studies have shown that biases in synthetic data can amplify disparities in AI predictions, especially in sensitive areas like healthcare. \cite{c11} conduct a scoping review on privacy and utility metrics for synthetic data, further elaborating on how these biases can undermine the effectiveness and trust of AI systems trained on synthetic medical data .

Synthetic data has also been explored in the specific context of brain tumor segmentation, a critical task in medical imaging. \cite{c5} present the BraTS 2015 dataset, which has become a standard benchmark for evaluating tumor segmentation algorithms, and subsequent works have used it to investigate the role of synthetic data in training segmentation models for brain tumors . \cite{c12} explore whether segmentation models can be trained effectively using fully synthetic data, finding that while fully synthetic datasets can perform similarly to real datasets, they often suffer from a generalization gap due to the synthetic data's limitations in capturing the complexity of real-world medical images . This work highlights the importance of understanding how the quality, diversity, and proportion of synthetic data impact the reliability of AI models, especially in specialized tasks like brain tumor segmentation.

\begin{figure*}[!ht]
    \centering
    \includegraphics[width=0.8\textwidth]{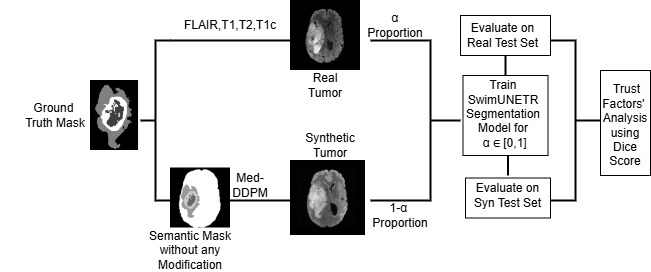} 
    \caption{Illustration of the Proposed Method}
    \label{fig:method_illustration}
\end{figure*}

\begin{figure*}[!ht]
    \centering
    \begin{subfigure}{0.16\textwidth}
        \includegraphics[width=\textwidth]{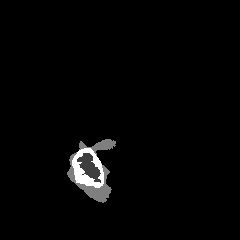}
        \caption{Mask}
        \label{fig:mask_real}
    \end{subfigure}
    \begin{subfigure}{0.16\textwidth}
        \includegraphics[width=\textwidth]{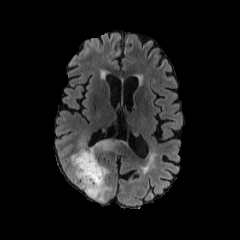}
        \caption{Flair (R)}
        \label{fig:flair_real}
    \end{subfigure}
    \begin{subfigure}{0.16\textwidth}
        \includegraphics[width=\textwidth]{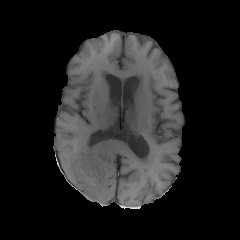}
        \caption{T1 (R)}
        \label{fig:t1_real}
    \end{subfigure}
    \begin{subfigure}{0.16\textwidth}
        \includegraphics[width=\textwidth]{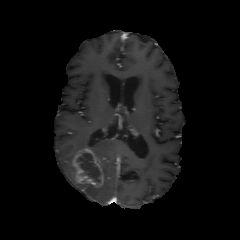}
        \caption{T1ce (R)}
        \label{fig:t1c_real}
    \end{subfigure}
    \begin{subfigure}{0.16\textwidth}
        \includegraphics[width=\textwidth]{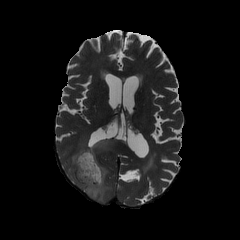}
        \caption{T2 (R)}
        \label{fig:t2_real}
    \end{subfigure}
    
    \vspace{0.5em} 

    \begin{subfigure}{0.16\textwidth}
        \includegraphics[width=\textwidth]{brats_20_mask_real.png}
        \caption{Mask}
        \label{fig:mask_synthetic}
    \end{subfigure}
    \begin{subfigure}{0.16\textwidth}
        \includegraphics[width=\textwidth]{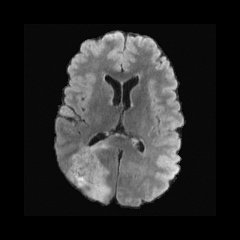}
        \caption{Flair (S)}
        \label{fig:flair_synthetic}
    \end{subfigure}
    \begin{subfigure}{0.16\textwidth}
        \includegraphics[width=\textwidth]{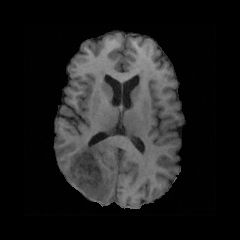}
        \caption{T1 (S)}
        \label{fig:t1_synthetic}
    \end{subfigure}
    \begin{subfigure}{0.16\textwidth}
        \includegraphics[width=\textwidth]{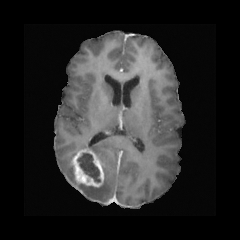}
        \caption{T1ce (S)}
        \label{fig:t1c_synthetic}
    \end{subfigure}
    \begin{subfigure}{0.16\textwidth}
        \includegraphics[width=\textwidth]{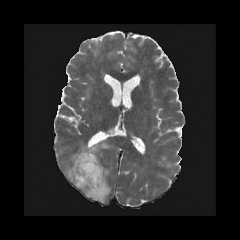}
        \caption{T2 (S)}
        \label{fig:t2_synthetic}
    \end{subfigure}

    \caption{Comparison of Real (R) and Synthetic (S) Brain Tumor Images.}
    \label{fig:brain_tumor_plot}
\end{figure*}

\section{EXPERIMENTAL SETUP}

Let the dataset be represented as:
$$
\mathcal{D} = \{(x_i, y_i)\}_{i=1}^N, \eqno{(1)}
$$
where $x_i$ denotes the multimodal MRI scans, $y_i$ represents the corresponding ground truth labels for tumor regions: Enhancing Tumor (ET), Peritumoral Edema (ED), and Necrotic Tumor Core (NCR), and $N$ represents the number of MRI samples. The dataset includes four modalities: FLAIR, T1, T2, and T1ce. 

The BraTS 2021 dataset \cite{c5}, comprising 1251 real multi-modal MRI images, is used. It is split into:
\begin{itemize}
    \item Real Training Set $\mathcal{D}_{R,T}$: 1000 images,
    \item Real Testing Set $\mathcal{D}_{R,t}$: 251 images.
\end{itemize}

A pre-trained Med-DDPM model provided by \cite{c6}, is used to generate the synthetic copies of the real MRI images, conditioned on the same semantic map without any modification. The generated synthetic dataset is:
$$
\mathcal{D}_{\text{syn}} = \{(x_i^{\text{syn}}, y_i)\}_{i=1}^N, \eqno{(2)}
$$
where the synthetic images $x_i^{\text{syn}}$ are created by using the same ground truth masks $y_i$ without any modifications or augmentations, ensuring one-to-one correspondence between real and synthetic data. So, the synthetic train and test sets can be defined as:
\begin{itemize}
    \item Synthetic Training Set $\mathcal{D}_{S,T}$: 1000 images,
    \item Synthetic Testing Set $\mathcal{D}_{S,t}$: 251 images.
\end{itemize}

Real MRI brain tumor images across all modalities, along with their synthetic counterparts, are presented in Fig. \ref{fig:brain_tumor_plot}. Notably, despite being generated using the same semantic map, subtle differences are observed between the real and synthetic samples. This experiment seeks to determine whether models trained on these datasets yield comparable results, thereby enabling an evaluation of the factors influencing trust in models utilizing synthetic data.

\section{METHODOLOGY}
\subsection{Segmentation Model}
A segmentation model $f_\theta(x)$, parameterized by $\theta$, is trained to predict the segmentation mask $\hat{y}_i$ for an input MRI scan $x_i$:
\[
\hat{y}_i = f_\theta(x_i), \eqno{(3)}
\]
where $f_\theta$ is implemented using the SwinUNETR\cite{c4} model, optimized for 3D semantic segmentation.

\subsection{Training and Testing Configuration}
The training dataset $\mathcal{D}_{\text{train}}$ for the segmentation, is constructed as a weighted combination of real and synthetic data:
\[
|\mathcal{D}_{\text{train}}| = \alpha \cdot |\mathcal{D}_{R,T}| + (1 - \alpha) \cdot |\mathcal{D}_{S,T}|, \eqno{(4)}
\]
where $\alpha \in [0, 1]$ controls the proportion of real data in the training set. By varying $\alpha$, different combinations of real and synthetic data are used to train the segmentation model.

The trained model is evaluated on the real test set $\mathcal{D}_{R,t}$ and the synthetic test set $\mathcal{D}_{S,t}$, each consisting of 251 MRI images.

\subsection{Evaluation Metric}
The performance of the segmentation model is evaluated using the Dice Similarity Coefficient (DSC), which measures the overlap between predicted ($\hat{y}_i$) and ground truth ($y_i$) segmentation masks:
\[
\text{Dice}(y_i, \hat{y}_i) = \frac{2 \cdot |y_i \cap \hat{y}_i|}{|y_i| + |\hat{y}_i|}. \eqno{(5)}
\]

\subsection{Analysis of Trust Factors}
Using the Dice scores, five key trust factors are quantitatively analyzed:
\paragraph{Overall Performance($T_1$)}
This factor captures the global quality of the model's predictions by computing the average Dice score across all tumor regions (ET, ED, NCR) for each test set (real and synthetic):

\[
T_{1,\beta} = \frac{1}{M}(\frac{1}{K} \sum_{k=1}^K \mathbf{C}_{\beta,k}), \quad \beta \in \{R, S\}. \eqno{(6)}
\]

Where:
\begin{itemize}
    \item \(T_{1,\beta}\) represents the average dice score for test set \(\beta\) (\(R\) for real and \(S\) for synthetic).
    \item $M=251$ represents the cardinality of test set.
    \item \(K = 3\), corresponding to the three tumor regions: \(ET\), \(ED\), and \(NCR\).
    \item \(\mathbf{C}_{\beta,k}\) is the dice vector for region \(k \in \{1, 2, 3\}\) (corresponding to \(ET\), \(ED\), and \(NCR\)) in test set \(\beta\).
\end{itemize}

By comparing \(T_{1,R}\) and \(T_{1,S}\), we can assess the alignment of model performance between real and synthetic datasets, which serves as a critical indicator of the synthetic data's reliability and effectiveness.

\paragraph{Consistency Across Regions (\(T_2\))}  
This factor measures the model’s ability to maintain consistent performance across different tumor regions (ET, ED, NCR). It is computed as follows:

\[
T_{2,\beta} = 1 - \text{std}(\frac{1}{K} \sum_{k=1}^K \mathbf{C}_{\beta,k}), \quad \beta \in \{R, S\}. \eqno{(7)}
\]

A higher \(T_2\) value indicates more consistent performance across the regions,implying that it does not favor any specific region over others and maintains balanced performance.

\paragraph{Alignment of Real and Synthetic Performance ($T_3$)} The alignment of model performance between the real and synthetic test sets is quantified using the cosine similarity between the Dice score vectors corresponding to each tumor region (ET, ED, NCR) in the real and synthetic test sets. It is given by:  

\[
T_{3,k} = \frac{\mathbf{C}_{R,k} \cdot \mathbf{C}_{S,k}}{\|\mathbf{C}_{R,k}\|  \|\mathbf{C}_{S,k}\|}  \eqno{(8)}
\]

A higher value of $T_3$ (closer to 1) indicates that the model's performance on the real data aligns closely with its performance on the synthetic data, suggesting that the synthetic data reliably reflects the characteristics of the real data. On the other hand, a lower cosine similarity may imply that the synthetic data does not capture the structure or distribution of the real data as effectively, which could affect the trustworthiness of the synthetic data.

\paragraph{Performance Difference ($T_4$)}
This factor quantifies the absolute difference in the performance of the model between the real and synthetic test sets across tumor regions (ET, ED, NCR). 

\[
T_{4,k} = \frac{1}{M} \sum_{i=1}^M |\mathbf{C}_{R,k} - \mathbf{C}_{S,k}|.  \eqno{(9)}
\]

 A lower $T_4$ value indicates that the model performs similarly on both real and synthetic data, suggesting that the synthetic data closely mimics the real data and can be trusted for model evaluation. Conversely, a higher $T_4$ value implies that the model's performance diverges between the two datasets, signaling that the synthetic data may not accurately reflect the characteristics of the real data, potentially reducing its reliability for clinical applications.

\paragraph{Predictive Variability ($T_5$)}
This factor quantifies the consistency of the model’s performance for a given tumor region by calculating the coefficient of variation (CV) of the Dice scores across both real and synthetic test sets. It is given by: 
\[
T_{5,\beta,k} = \frac{\text{std}(\mathbf{C}_{\beta,k})}{\text{mean}(\mathbf{C}_{\beta,k})}   \eqno{(10)}
\]

A lower value of $T_5$ for the synthetic test set indicates that the model’s performance remains consistent and that the synthetic data effectively captures real-world variability. Conversely, a higher $T_5$ value reflects greater instability, suggesting that the synthetic data might not reliably represent real-world conditions.
 
\section{RESULTS AND DISCUSSIONS}
\subsection{Analysis of Tumor Regions}

\begin{figure}[!ht]
    \centering
    \begin{subfigure}{\columnwidth}
        \centering
        \includegraphics[width=0.8 \textwidth]{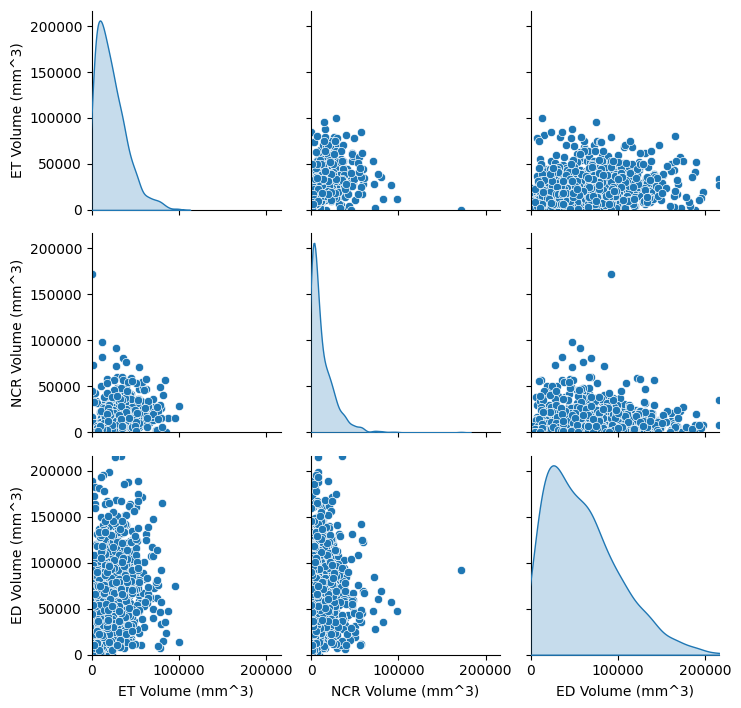}
        \caption{Training Pair Plot}
        \label{fig:pair_plot_training}
    \end{subfigure}
    
    \vspace{0.5em} 
      
    \begin{subfigure}{0.8\columnwidth}
        \centering
        \includegraphics[width=\textwidth]{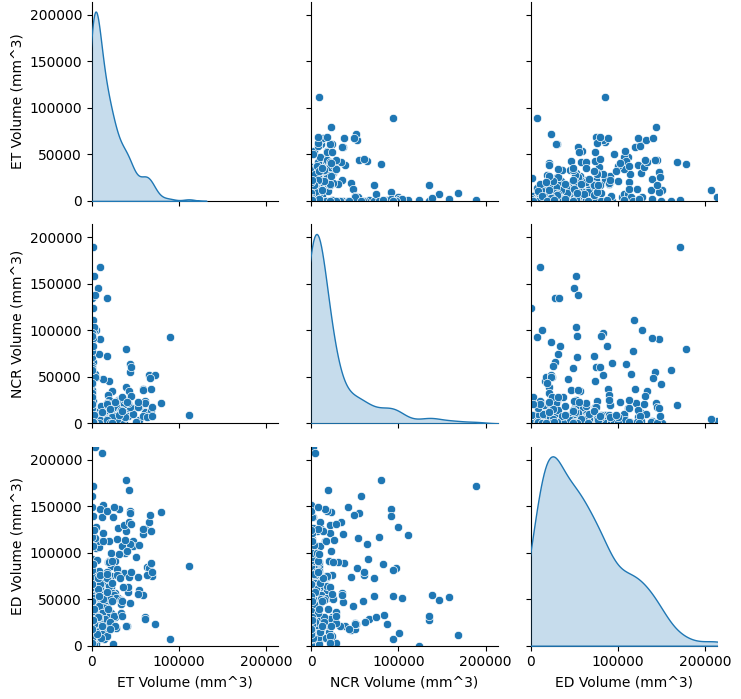}
        \caption{Testing Pair Plot}
        \label{fig:pair_plot_testing}
    \end{subfigure}
    
    \caption{Volumetric Distribution of Tumor Regions}
    \label{fig:pair_plots_vertical}
\end{figure}

The volume of each tumor region (ET, ED, NCR) is calculated by counting the number of voxels occupied by the respective region in the segmentation mask. The volumetric distribution of these regions in the training and testing sets is illustrated in Fig. \ref{fig:pair_plots_vertical}. The plot highlights a significant class imbalance in the training data, with ED occupying the largest volume, followed by ET, and NCR being the smallest. A similar distribution is maintained in both the real and synthetic test sets to enable an effective evaluation of the proposed trust factors.

\subsection{Trust Factor \texorpdfstring{$T_1$}{T\_1}}
\begin{figure}[!ht]
    \centering
    \begin{subfigure}{\columnwidth}
        \centering
        \includegraphics[width=0.85\textwidth]{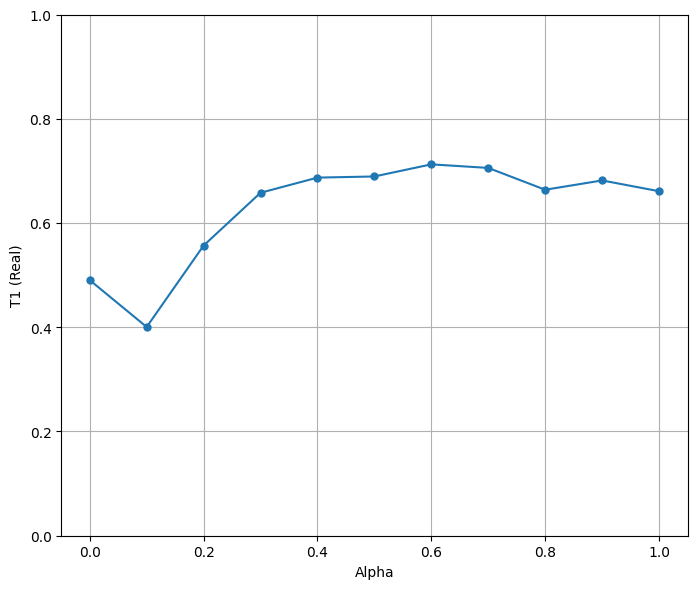}
        \caption{Real Test Set}
    \end{subfigure}
    \vspace{0.5em}
    \begin{subfigure}{\columnwidth}
        \centering
        \includegraphics[width=0.85\textwidth]{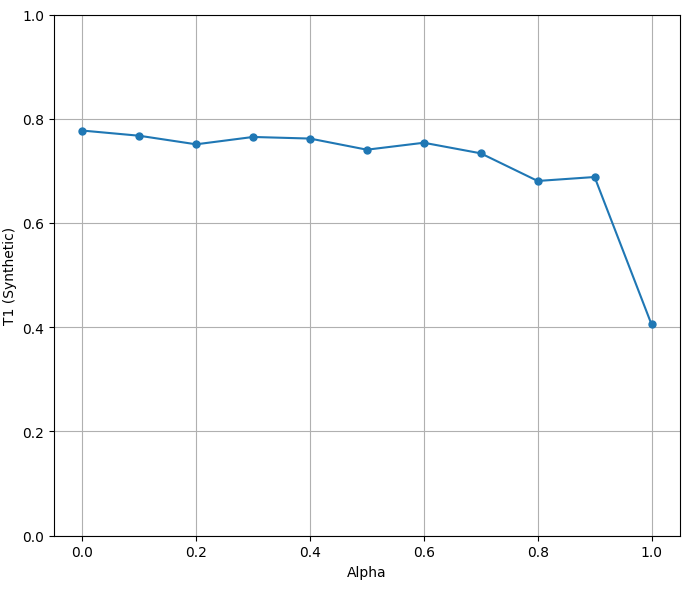}
        \caption{Synthetic Test Set}
    \end{subfigure}
    
    \caption{Visualization of Trust Factor $T_1$}
    \label{fig:t1}
\end{figure}

Theoretically, the values of $T_{1,R}$ and $T_{1,S}$ should be nearly identical for $\alpha \in [0,1]$, as the synthetic data is merely a replica of the real MRI. However, as seen in Fig. \ref{fig:t1}, a significant discrepancy is observed in the $T_1$ values at the extreme values of $\alpha=0$ (trained on all synthetic images) and $\alpha=1$ (trained on all real images). This highlights the importance of using an appropriate combination of real and synthetic data during training to obtain reliable results. Notably, at $\alpha=0.5 \pm 0.1$, representing a balanced proportion, the $T_1$ values are quite similar, as also emphasized in \cite{c8}. This leads to the important conclusion that a balanced mix of real and synthetic datasets is crucial to enhancing trust in clinical AI. 

\subsection{Trust Factor \texorpdfstring{$T_2$}{T\_2}}
\begin{figure}[!ht]
    \centering
    \begin{subfigure}{\columnwidth}
        \centering
        \includegraphics[width=0.85\textwidth]{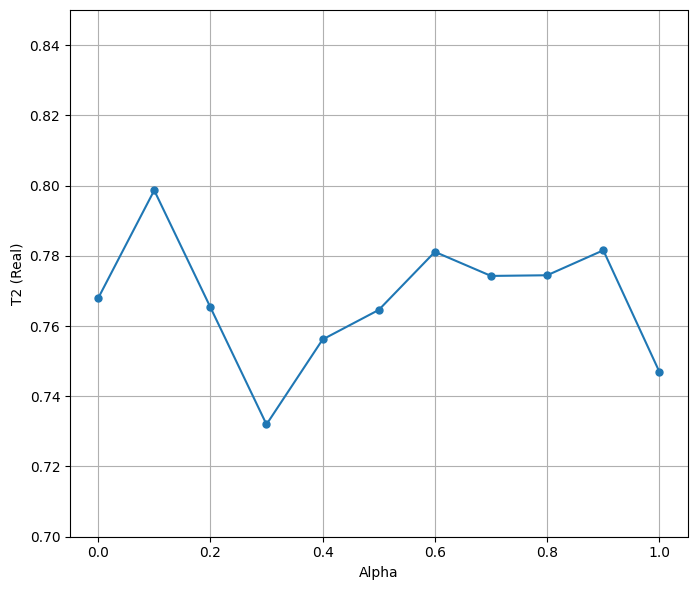}
        \caption{Real Test Set}
    \end{subfigure}
    \vspace{0.5em}
    \begin{subfigure}{\columnwidth}
        \centering
        \includegraphics[width=0.85\textwidth]{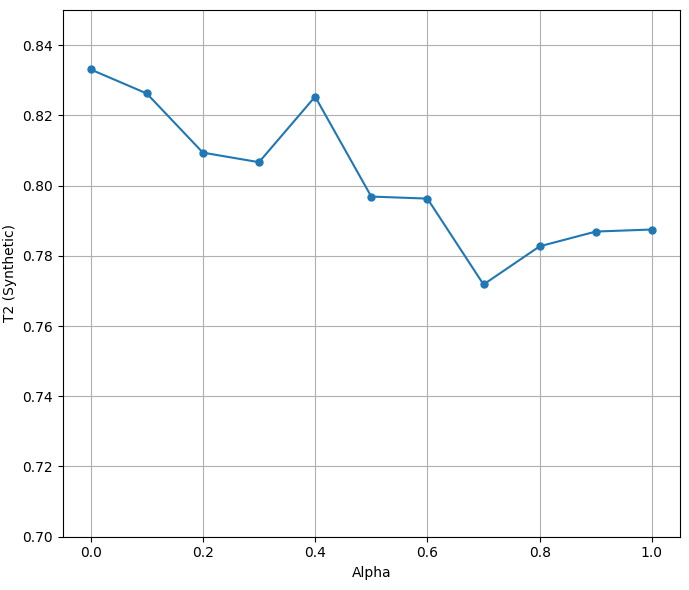}
        \caption{Synthetic Test Set}
    \end{subfigure}
    
    \caption{Visualization of Trust Factor $T_2$}
    \label{fig:t2}
\end{figure}

Consistency in segmentation performance across all regions, regardless of their size, is a critical factor for building trust. As shown in Fig. \ref{fig:t2}, $T_{2,S}$ is consistently higher than $T_{2,R}$ for almost every $\alpha$, indicating that the performance on synthetic data is fairer and more consistent compared to real data. Although the DDPM-generated synthetic data is conditioned on the same semantic maps, it does not fully capture the inherent biases and variability present in the real data. This highlights the importance of utilizing high-quality real and synthetic data, free from such biases and imbalances, to ensure reliable results and enhance trust in synthetic data for clinical AI applications.

\subsection{Trust Factor \texorpdfstring{$T_3$}{T\_3}}
\begin{figure}[!ht]
    \centering
    \includegraphics[width=0.48\textwidth]{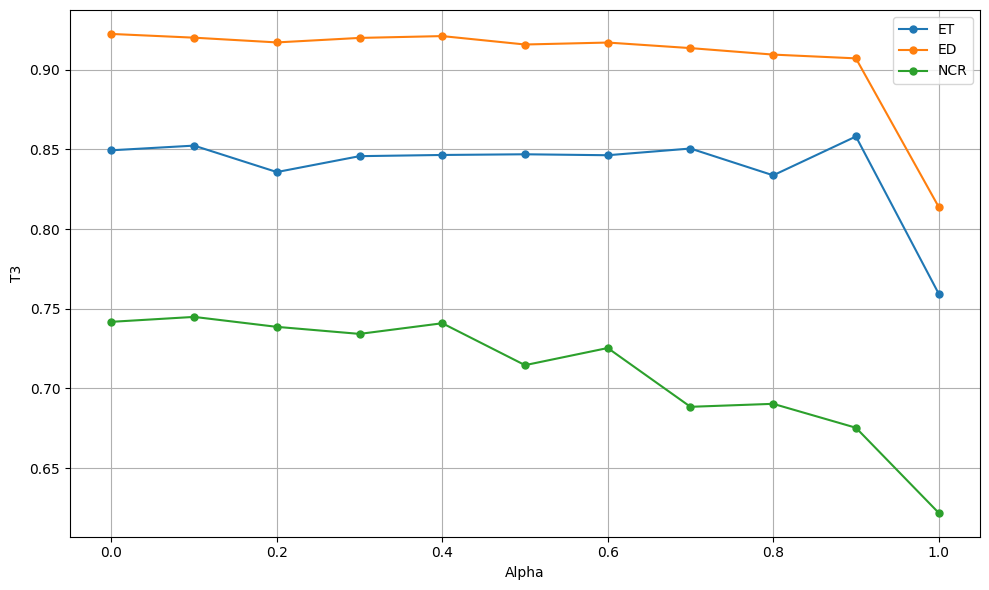} 
    \caption{Visualization of Trust Factor $T_3$}
    \label{fig:t3}
\end{figure}

As shown in Fig. \ref{fig:t3}, which illustrates the alignment of Dice scores between the real and synthetic test sets, the $T_3$ value for ED is significantly higher than that for ET, followed by NCR, across all values of $\alpha$. This trend arises because ED occupies a substantially larger volume in the dataset compared to ET and NCR, as depicted in Fig. \ref{fig:pair_plots_vertical}. Additionally, for all three regions (ET, ED, and NCR), there is a noticeable downward trend in $T_3$ values from $\alpha = 0$ to $\alpha = 1$, indicating that bias is more prominent in the real data compared to the synthetic data, a finding consistent with $T_2$. 

Notably, the decline in $T_3$ is most pronounced for NCR, which has the smallest volume. These observations suggest that the reliability and trustworthiness of synthetic data are directly proportional to the size of the attributes. This underscores a critical limitation: synthetic data may not be as trustworthy for tasks involving regions of interest that are quite small, such as microscopic structures.

\subsection{Trust Factor \texorpdfstring{$T_4$}{T\_4}}
\begin{figure}[!h]
    \centering
    \includegraphics[width=0.48\textwidth]{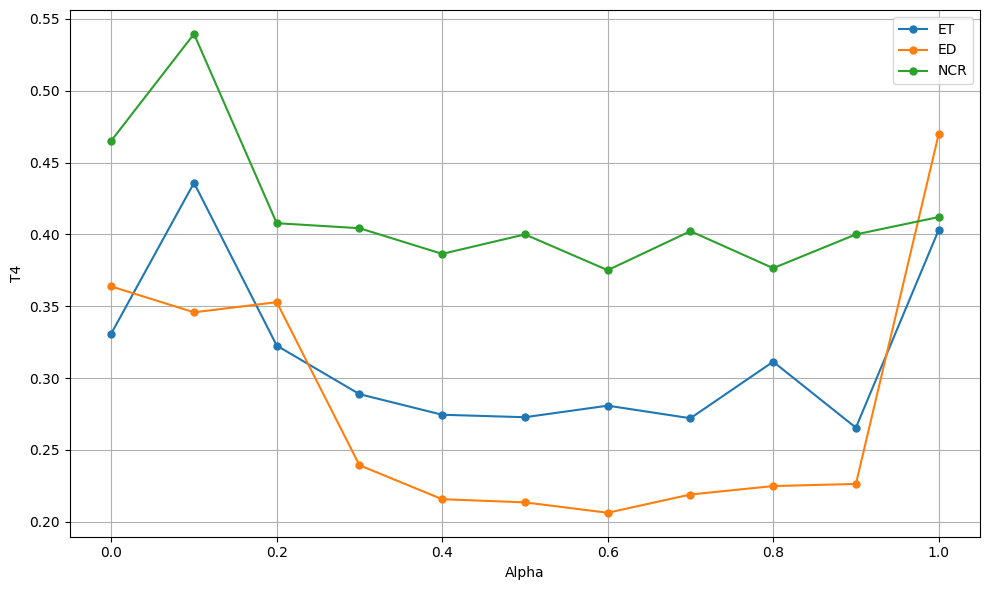} 
    \caption{Visualization of Trust Factor $T_4$}
    \label{fig:t4}
\end{figure}

The mean absolute difference in performance, as shown in Fig. \ref{fig:t4}, reveals a pattern closely tied to the volume of the tumor regions, with $T_4$ values being significantly lower for ED compared to ET and NCR. When the model is trained entirely on synthetic data ($\alpha = 0$), the $T_4$ value is highest for NCR, followed by ED, and lowest for ET. As the training transitions to real data ($\alpha = 1$), the $T_4$ value decreases for NCR but increases significantly for ED.  

These results emphasize that synthetic data struggles to reliably and consistently capture the characteristics of smaller regions like NCR, a limitation also evident in the $T_3$ analysis. A noteworthy observation is that the minimum $T_4$ value for all three tumor regions occurs at $\alpha = 0.5 \pm 0.1$, aligning with the findings from $T_1$.  

This highlights the importance of incorporating greater diversity in synthetic data, especially for underrepresented regions. Strategies such as targeted augmentation or advanced conditioning mechanisms could help address this imbalance, ensuring more robust and reliable performance across all regions. 

\subsection{Trust Factor \texorpdfstring{$T_5$}{T\_5}}
\begin{figure}[!ht]
    \centering
    \begin{subfigure}{\columnwidth}
        \centering
        \includegraphics[width=\textwidth]{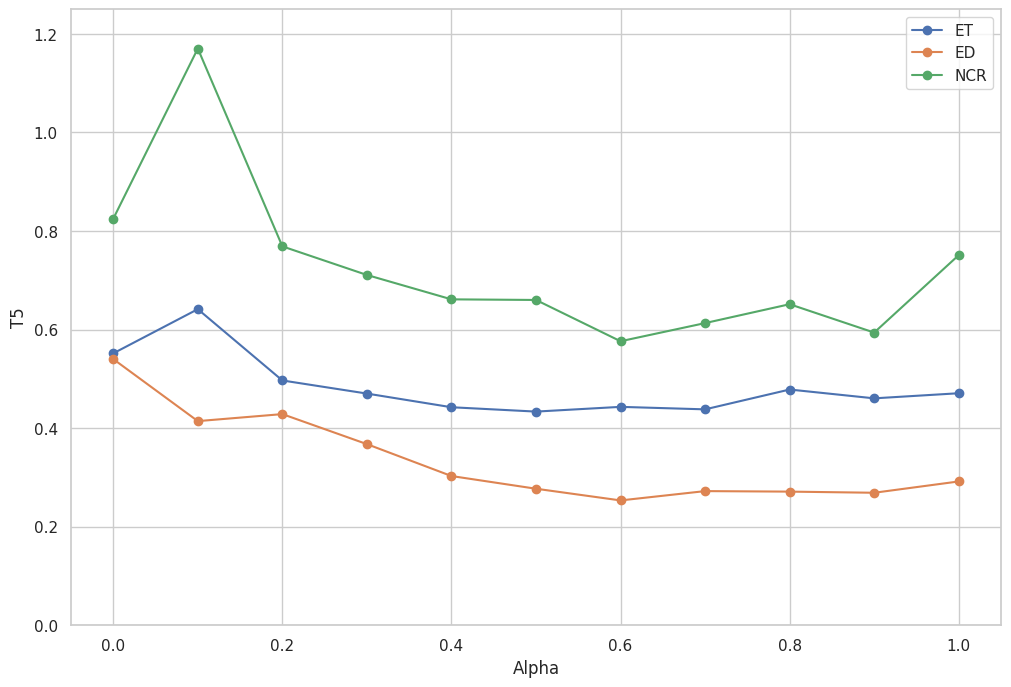}
        \caption{Real Test Set}
    \end{subfigure}
    \vspace{0.5em}
    \begin{subfigure}{\columnwidth}
        \centering
        \includegraphics[width=\textwidth]{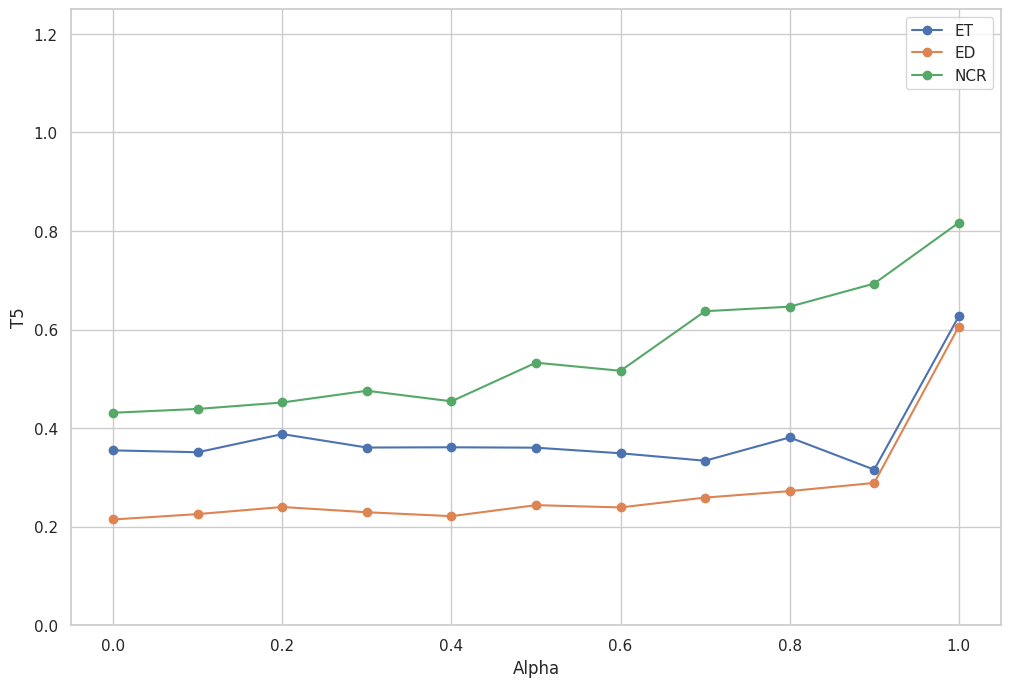}
        \caption{Synthetic Test Set}
    \end{subfigure}
    
    \caption{Visualization of Trust Factor $T_5$}
    \label{fig:t5}
\end{figure}

From Fig. \ref{fig:t5}, it is noteworthy that $T_{5,S}$ is significantly lower than $T_{5,R}$ for all tumor regions across $\alpha \in [0,1]$. Moreover, the behavior of $T_{5,S}$ is relatively stable for $\alpha \in [0.5 \pm 0.2]$, whereas $T_{5,R}$ exhibits high variability across all values of $\alpha$. This suggests that synthetic data lacks the ability to fully capture the nuances present in real data. To address this limitation, it is essential to ensure the use of high-quality real and synthetic data through appropriate preprocessing techniques, as also highlighted in the case of $T_2$.

Additionally, examining the extreme cases of $\alpha=0$ (model trained entirely on synthetic replicas) and $\alpha=1$ (model trained entirely on real samples), a similar pattern is observed for both $T_{5,R}$ and $T_{5,S}$. Specifically, the model trained solely on real data struggles to generalize to synthetic data and vice versa, despite the synthetic replicas being designed to exhibit similar characteristics. This emphasizes the importance of leveraging advanced generative models to produce synthetic images that accurately capture fine-grained details.

Interestingly, the $T_{5,R}$ and $T_{5,S}$ values for ED are quite comparable, indicating that synthetic data performance can be trusted for larger tumor regions, as seen in the case of $T_3$. Visual inspection of the samples in Fig. \ref{fig:brain_tumor_plot} further reinforces this observation. For instance, in the FLAIR (Real) images, the NCR region is distinctly visible, whereas in the FLAIR (Synthetic) images, only the ED region, which occupies the largest outer volume, is prominently visible.

This highlights the necessity of training models using a selectively curated dataset that includes well-delineated regions with substantial volumes. Doing so would enhance the trustworthiness of synthetic data in clinical AI applications, ensuring it better represents the nuances of real-world variability. 

\section{CONCLUSION}

Synthetic data holds transformative potential for advancing artificial intelligence in healthcare, but its successful integration hinges on building trust among clinicians and healthcare professionals. Our discussion emphasizes that while synthetic data can alleviate issues such as data scarcity and bias, its widespread adoption is impeded by concerns over its reliability and clinical relevance. To overcome these barriers, it is essential to focus on the quality (representativeness), diversity (size of attributes), and proportion (number of images) of synthetic datasets, ensuring they mirror real-world complexities. Moreover, fostering transparency and explainability in synthetic data generation processes will be key to addressing the trust deficit. As the healthcare industry moves toward adopting AI solutions, we must prioritize research that strengthens the reliability of synthetic data, ensuring it complements, rather than replaces, real data. Only then can we unlock the full potential of synthetic data to improve patient outcomes, reduce healthcare disparities, and create more robust, trustworthy AI systems for clinical practice.

.

\end{document}